\documentclass[10pt, logo, onecolumn, copyright]{nvidiatechreport}

\usepackage[ruled,vlined]{algorithm2e}
\usepackage[numbers]{natbib}
\usepackage{multirow}
\usepackage{xspace}
\usepackage{enumitem}
\usepackage{pifont}
\usepackage{subcaption}
\usepackage[most]{tcolorbox}
\usepackage[table,xcdraw]{xcolor}

\newtcolorbox{theorybox}[1][]{
  enhanced, breakable=false,
  colback=gray!4, colframe=black!70,
  boxrule=0.5pt, arc=1pt,
  left=5pt, right=5pt, top=3pt, bottom=3pt,
  title={#1}, fonttitle=\bfseries\small,
  coltitle=black, colbacktitle=gray!12,
  before skip=6pt, after skip=6pt}
\newcommand{\act}{\textsf{act}\xspace}
\newcommand{\abstain}{\textsf{abstain}\xspace}
\newcommand{\xA}{\ensuremath{x_{\mathrm{Audio}}}\xspace}   
\newcommand{\xT}{\ensuremath{x_{\mathrm{Text}}}\xspace}   
\newcommand{\yA}{\ensuremath{y_{\mathrm{Audio}}}\xspace}   
\newcommand{\yT}{\ensuremath{y_{\mathrm{Text}}}\xspace}   
\newcommand{\lt}{\textsc{LT}\xspace}                    

\makeatletter\renewcommand\paragraph{\@startsection{paragraph}{4}{\z@}
  {.5em \@plus1ex \@minus.2ex}{-.5em}{\normalfont\normalsize\bfseries}}\makeatother

\newcommand{\vm}{\textsc{Voice Memory}\xspace}
\newcommand{\rir}{\ensuremath{\rho}\xspace}
\newcommand{\her}{\textsc{her}\xspace}

\newcommand{\yes}{\textcolor{green!60!black}{\ding{51}}}   
\newcommand{\no}{\textcolor{gray!80!black}{\ding{55}}}      

\title{Voice Memory for Agentic Speech Recognition}

\author{
Chao-Han Huck Yang$^{*}$ ~
Zih-Ching Chen ~
Piotr \.{Z}elasko ~
Zhehuai Chen ~
Jagadeesh Balam ~
Boris Ginsburg \\
\vspace{2mm}
{\normalsize NVIDIA} \\
\vspace{2mm}
{\normalsize Correspondence: \texttt{huckiyang@ieee.org} ~ \texttt{\{virginiac, pzelasko, zhehuaic\}@nvidia.com}\\$^{*}$Work done at NVIDIA. } \\
\vspace{2mm}
{\normalsize
\href{https://huggingface.co/huckiyang/voice-memory}{Model} ~~~~
\href{https://huckiyang.github.io/voice-memory/}{Demo} ~~~~
\href{https://github.com/NVIDIA-NeMo/Speech}{NeMo-Speech} ~~~~
\href{https://github.com/NVIDIA/NemoClaw}{NeMoClaw} ~~~~
\href{https://huckiyang.github.io/voice-memory-notebook}{Notebook}
}
}

\keywords{voice agents, memory modeling, on-device, language models, test-time adaptation}

\begin{abstract}
\textbf{Abstract}\\
We present \vm{}, a inference-only scheme for agentic speech recognition: at stream time, a frozen corrector reads a single per-domain \texttt{memory.md} and decides per utterance whether to \emph{act} on the hypothesis or \emph{abstain} and keep the 1-best. Asynchronously, a score-gated optimizer revises that file through bounded edits, accepting an edit only when it strictly improves a held-out score. Extended from classical ASR-LM framework, we refer this split the \emph{listener-thinker} architecture; the two roles are coupled only through the memory, so no weights change and the learned skill stays auditable and portable. Restraint turns out to be the operative skill this loop discovers: unconstrained generative error correction (GER) over-corrects, breaking correct tokens on up to 64\% of its edits on financial news, and \vm{} reduces this rate to 35\%. Across ten HyPoradise domains with an open corrector, \vm{} lowers weighted word error rate from 8.36\% to 7.52\% (7.47\% with three added in-context examples) without regressing any dataset below its 1-best baseline; gains concentrate where recoverable headroom is largest, including air-travel commands (8.40\% to 3.40\%) and noisy far-field speech (CHiME-4, 12.69\% to 10.46\%). The memory transfers across corrector families and adds zero parameters to the inference path. A demo and example code are provided for future studies. 
\end{abstract}

\begin{document}
\maketitle

\section{Introduction}
\label{sec:intro}
Modern voice systems have entered a low-error regime for clean speech recognition. Decades of progress, from statistical HMM systems \cite{rabiner1989tutorial,jelinek1997statistical} to large-scale neural models \cite{baevski2020wav2vec,hsu2021hubert,gulati2020conformer,radford2023robust,puvvada2024less}, have brought read speech below 2\% word error rate (WER); yet decoders still make systematic, domain-specific errors \cite{chen2023hyporadise,likhomanenko2021rethinking,szymanski2020wer,ma2023n} and preferences~\cite{koluguri2026preference,hu2025chain, wu2026speecheq, hu2023scaling}. In the classical source-channel formulation of recognition \cite{jelinek1997statistical}, which mediates between words and memories~\cite{goldinger1996words, papcun1989long}, these are exactly the failures expected when acoustic evidence is reconciled with a mismatched language prior. These errors~\cite{szymanski2020wer,stolcke2000dialog} are consistent within a domain but differ across domains: in financial news, an ASR system renders ``\texttt{Bentsen}'' as ``\texttt{benson}''; in air-travel queries, it collapses the spelled form ``\texttt{u s air}'' into ``\texttt{us air}.'' A fix tuned for one domain therefore transfers poorly to another.

\begin{figure}[t]
    \centering
    \includegraphics[width=0.83\linewidth]{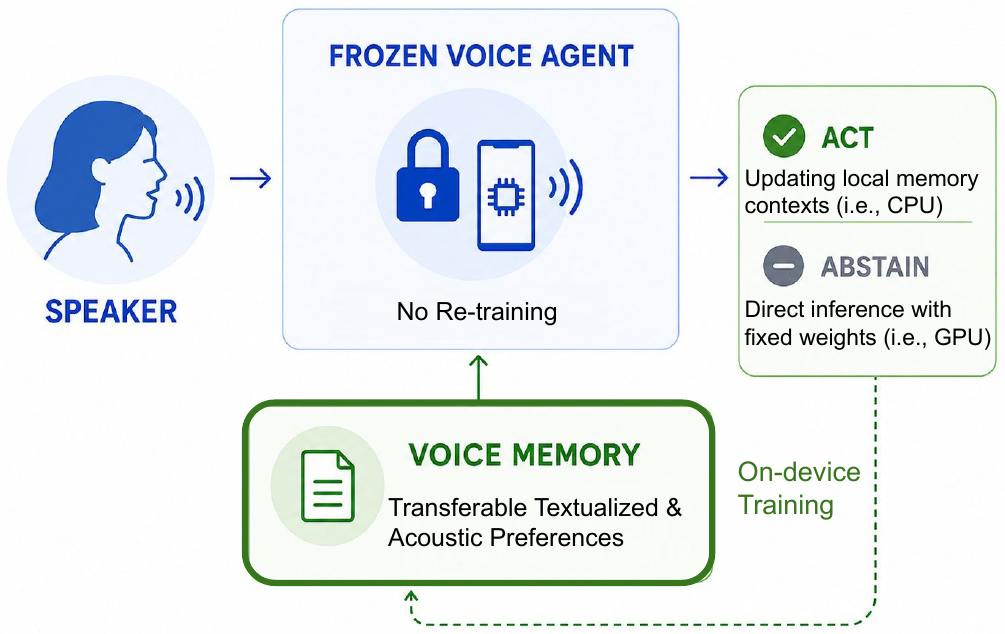}
\caption{Overview of \vm{}. At inference, a frozen corrector reads a learnable,
human-readable text memory and decides per utterance whether to \emph{act} on
the ASR hypothesis or \emph{abstain}. Offline, an optimizer distills this
instructional memory from training corpora using forward passes only, with no
weight updates, so that the learned skill (i) generalizes within an acoustic
domain and (ii) transfers across domains and correctors.}
    \label{fig:fig1}
\end{figure}

A widely adopted remedy is the generative hypothesis-rewriting framework: the $n$-best decoding hypotheses (\textit{i.e.}, ASR, OCR, or machine translation) are passed to an LLM, which rewrites them into a single transcript \cite{chen2023hyporadise,radhakrishnan2023whispering,thomas2024leveraging,yeo2025mms,hsu2025let, yang2025covoger,carofilis2026text}. For instance, GER~\cite{yang2023generative} extends a long line of language-model rescoring and correction, from BERT-based hypothesis scoring \cite{salazar2020masked} to edit-aligned neural correction \cite{leng2021fastcorrect} that achieved several state-of-the-art results in annual speech-translation competitions and on robust ASR benchmarks. In the low-error regime, however, a strong LLM tends to \emph{over-correct}: it rewrites tokens that were already correct, changing \texttt{home is} to \texttt{home's} and normalizing digits and spelling to its own conventions. Each such edit is fluent, yet it moves the output away from the reference transcript; on low-headroom domains, zero-shot GER raises WER above the unedited 1-best (\S\ref{sec:scaling}). The decision facing a corrector has therefore inverted: at sub-2\% WER, the question is no longer whether the model \emph{can} fix an error, but whether it should modify the hypothesis at all.

We refer to this setting as \emph{agentic speech recognition}, and we make the term precise. We call an ASR system agentic when it satisfies three conditions. First, \textbf{decision}: it selects among actions per utterance, in our case whether to \emph{act} on the hypothesis or to \emph{abstain} and keep it, rather than applying one fixed transformation to every input. Second, \textbf{persistent state}: its decisions are conditioned on an explicit, inspectable state that outlives a single utterance, in our case a text memory rather than activations or weights. Third, \textbf{self-improvement}: it revises that state from feedback on its own scored outputs, without human intervention. Prior work realizes these properties inside the model, fine-tuning an omni-model to emit per-utterance action decisions from audio (\S\ref{sec:related}). We study the complementary, test-time realization: the decision policy lives in an external memory that a frozen corrector reads at inference.

Definition~1 collects the three conditions into a single format, which we call the \emph{listener-thinker} (\lt{}) architecture. It extends the generic ASR-LM cascade, a frozen decoder ($B_\phi$) feeding a frozen language model $M_\theta$, with exactly the two objects the three conditions require: an explicit action distribution $\pi$ over $\{\act, \abstain\}$ and a persistent text state ($s$) that a separate, asynchronous optimizer revises. The listener is the synchronous path that decodes and decides at stream time; the thinker is the background path that improves $s$ between utterances. Table~\ref{tab:typology} situates the format against the two dominant speech stacks: E2E ASR fixes the map and never decides, SpeechLMs always emit and keep their state in weights and context, and only the \lt{} format has slots for evidence ($H$), external state, a per-utterance action $a$, and self-improvement.

\begin{theorybox}[Definition 1: The Listener-Thinker (\lt) Format]
\small
Write $x = (\xA, \xT)$ for audio input and $\hat{y} = (\yT, \yA)$ for 
text output. An agentic speech system in the \emph{listener-thinker} is a
tuple $(B_\phi, M_\theta, \pi, s, Op)$, whose synchronous \emph{listener}
$(B_\phi, M_\theta, \pi)$ works with an optimizer ($Op$): a frozen decoder $B_\phi$ maps $x$ to
textual evidence $H = (h_1, \dots, h_n)$; a frozen $M_\theta$, holding a
persistent, inspectable state $s \in \Sigma^{*}$, emits per utterance through
an action $a \sim \pi(\cdot \mid H, s)$, so that
$p(\hat{y} \mid x, s) = \sum_{a} \pi(a \mid H, s)\, p_\theta(\hat{y} \mid H, s, a)$
with:
\begin{equation*}
\hat{y} \;=\;
\begin{cases}
M_\theta(H, s) & a = \act\\[1pt]
h_1            & a = \abstain,
\end{cases}
\end{equation*}
while text state $s$ is revised only asynchronously by the score-gated
\emph{thinker}, with $\theta, \phi$ fixed.
\textsc{asr-lm} (this work) and \textsc{speechlm-lm} are the same type at
different signatures: $\xT = \yA$ at utterance timescale versus
the full signature at frame timescale with
$\mathcal{A} \supseteq \{\act, \abstain\}$ (speak, wait, delegate). Constant
policies recover the classics: $\pi \equiv \abstain$ is E2E decoding;
$\pi \equiv \act$, $s = \varnothing$ is a zero-shot GER case, when the thinker is with fixed $\theta$ and $ \phi$ parameters with failure rollouts ($F$) and Success rollouts ($U$).
\label{def:1}
\end{theorybox}

\noindent\textbf{Remark 1 (Relation to interaction models).}
The \lt{} format is the speech instance of the interaction/background split
now emerging at the frontier~\cite{tml2026interaction,huang2026duplexomni}:
the listener is the synchronous interaction path, the thinker is the
asynchronous background path, and the two are coupled \emph{only} through the
persistent artifact text state, $s$. The artifact is what makes the pair auditable and
portable.

\begin{table}[t]
\caption{Typology of speech systems. The listener-thinker (\lt) format
subsumes \textsc{asr-lm} (refer to Definition~\ref{def:1}, utterance timescale) and \textsc{speechlm-lm} (full signature, frame timescale) as one
type: only the \lt{} column has slots for evidence $H$, external text state $s$,
a per-utterance action $a$, and test-time self-improvement.}
\label{tab:typology}
\centering
\small
\setlength{\tabcolsep}{6pt}
\renewcommand{\arraystretch}{1.35}
\begin{tabular}{@{}lccc@{}}
\toprule
 & \textbf{E2E ASR} & \textbf{SpeechLM} & \textbf{Listener-Thinker (agentic; cascaded)} \\
\midrule
Distribution & $p_\phi(\yT \,{\mid}\, \xA)$
             & $p_\theta(\yT, \yA \,{\mid}\, \xT, \xA)$
             & $\pi(a \,{\mid}\, H, s)\, p_\theta(\hat{y} \,{\mid}\, H, s, a)$ \\
Evidence     & \xA
             & $\xA, \xT$
             & $H \sim B_\phi(x)$ \\
State        & weights $\phi$
             & weights $+$ contexts.
             & external text $s \in \Sigma^{*}$ \\
Decision     & none (fixed map)
             & none (always emit)
             & $\mathcal{A} \supseteq \{\act, \abstain\}$ \\
Update       & offline $\nabla_\phi$
             & offline $\nabla_\theta$ / ICL
             & $s' = Op(s, F, U)$, $v\!\uparrow$ async \\
Objective    & WER
             & likelihood
             & verifiable rewards vs.\ oracle \\
Instances    & Whisper~\cite{radford2023robust}, Parakeet~\cite{rekesh2023fast}
             & Moshi~\cite{defossez2024moshi}, SALMONN~\cite{tang2024salmonn}
             & ASR-LMs~\cite{yang2023generative}, DuplexLM-LMs~\cite{thinkingmachines2026interactionmodels}\\
\bottomrule
\end{tabular}
\end{table}

Expressing such a policy with existing adaptation mechanisms is costly. Fine-tuning, LoRA \cite{hu2022lora}, and soft prompts \cite{lester2021power} place the adaptation in weights or continuous vectors, which cannot be inspected, transferred across model families, or revised without retraining; few-shot exemplars \cite{brown2020language} recur a token cost on every request. None yields an adaptation that is simultaneously inference-only, auditable, and portable across correctors.

We present \vm{}, which stores the correction skill in a text file. The corrector stays frozen and reads a single human-readable \texttt{memory.md} at test time (Figure~\ref{fig:fig1}). A separate optimizer converts scored rollouts into bounded add/delete/replace edits and accepts an edit only when it strictly improves a held-out score \cite{yang2026skillopt}. The dominant policy this loop discovers is restraint: from its own scored mistakes, the optimizer writes suppressive rules such as ``\texttt{keep the verbatim ASR token; do not normalize},'' with no human in the loop. Where recoverable headroom exists, \vm{} closes most of the gap between the 1-best and the oracle;  a second corrector running the same loop even surpasses the $n$-best oracle by recovering tokens present in no hypothesis. We will later introduce new measurement metrics of (i) recoverable information ratio ($
\rho$) and (ii) harmful edit rate on these two scenarios for agentic speech recognition. 

This paper makes four contributions:
\begin{enumerate}[leftmargin=1.4em,itemsep=2pt,topsep=2pt]
\item \textbf{Method.} \vm{}, an inference-only adaptation scheme for agentic speech recognition: the correction skill resides in a per-domain text memory that a validation-gated loop improves through bounded edits, with no weight updates (\S\ref{sec:method}).
\item \textbf{Formulation and diagnostics.} A formal account of the act/abstain decision as the listener-thinker format (Definition~1, Table~\ref{tab:typology}), together with two measures: the \emph{Recoverable Information Ratio} (\rir{}) for recovery against the $n$-best oracle, and the \emph{Harmful Edit Rate} (\her{}, Figure~\ref{fig:her}) for over-correction. Across nine domains, the gain from correction tracks recoverable headroom ($r{=}{+}0.90$), delineating the operating regime in which a corrector should act (\S\ref{sec:scaling}).
\item \textbf{Portability.} Evidence that the optimized memory is a transferable, corrector-agnostic artifact: a memory written by one model family improves a reader from another, and that reader can re-form its own with the same loop (\S\ref{sec:transfer}).
\item \textbf{Noise robustness.} The same restraint policy holds under acoustic noise on CHiME-4 and NOIZEUS, matching training-based robust correction without fine-tuning or latent noise embeddings (\S\ref{sec:robust}).
\end{enumerate}

Taken together, \vm{} offers a low-cost adaptation path for cascades of ASR and LLM correction: no training infrastructure, a sub-10\,KB auditable artifact, and a measured account of when a corrector should act and when it should hold still.

\section{Related Work}
\label{sec:related}

\paragraph{ASR-LM and Agentic Setup for ASR.}
Recent work treats audio LLMs as agents that decide how to use internal perception and external evidence, rather than decoding passively \cite{wan2026speechhands,lin2025neko}. Speech-Hands \cite{wan2026speechhands} and Speech-Mind~\cite{wang2026audio} are the closest to our work. These setups learn a per-utterance action token by fine-tuning an omni-model on speech and audio reasoning, so its agency is acoustic, learned in weights, and re-decided each utterance. \vm{} differs on each of these axes (\S\ref{sec:intro}): it learns one accumulated memory, uses no audio and no training, and runs entirely at inference. The two methods are the trained and test-time versions of one idea, an ASR system that decides when to trust its own output.

\paragraph{Generative error correction and contextual biasing.}
Injecting domain-specific knowledge into ASR output is a long-standing goal. Early
systems brought contextual $n$-grams into decoding through shallow fusion and on-the-fly
rescoring \cite{aleksic2015bringing,zhao2019shallow}. End-to-end systems then moved
biasing into the model: CLAS \cite{pundak2018deep} jointly embeds bias phrases and
attends to them during decoding, and later work extended this to RNN-T
\cite{jain2020contextual} and to trie-based deep biasing \cite{le2021contextualized}.
These methods adapt the acoustic model and require a bias list and a training run.
LLM-based GER instead recasts correction as a language task over $n$-best hypotheses
\cite{chen2023hyporadise,yang2023generative,radhakrishnan2023whispering}, adapting the
corrector through fine-tuning or in-context exemplars. \vm{} sits at the opposite end of
both lines: it leaves the corrector frozen and writes the domain's conventions into a
text memory read at inference, with no bias list, no training, and no weight updates.

\paragraph{Noise-robust ASR and robust hypotheses.}
One popular ASR-LM use case is Robust-ASR. Acoustic noise makes correction harder because the $n$-best list itself degrades, so a corrector must denoise rather than rerank. The Robust HyPoradise benchmark \cite{hu2024large} pairs noisy Whisper hypotheses with clean references across CHiME-4, VoiceBank-DEMAND, LibriSpeech-FreeSound, and NOIZEUS. RobustGER distills the diversity of those hypotheses into a language-space noise embedding and fine-tunes the corrector against it, which recovers much of the lost accuracy. Cross-modality correctors such as NeKo \cite{lin2025neko} extend this with task-guided mixtures of experts. Both methods read the noise condition in a latent vector and pay for it with a training run. \vm{} uses the same observation, that the $n$-best spread already encodes the noise condition, but writes the prior into a per-noise-family text memory that the frozen corrector reads at inference (\S\ref{sec:robust}).

\paragraph{Text-space optimization.}
Optimizing natural-language artifacts such as prompts and skills against a score is an alternative to gradient updates \cite{yuksekgonul2024textgrad, yang2024large, yang2026skillopt}.
We apply this to ASR correction, where the $n$-best oracle sets a bounded target (\rir{}) and the main learned behavior is restraint.

\paragraph{WER critique and semantics.}
Prior work questions WER as the only ASR metric \cite{szymanski2020wer,kim2021semantic}. We add two results. We measure the gap between surface error and meaning (\S\ref{sec:relativity}), and we show that a semantic training signal gives lower WER than a traditional WER signal (\S\ref{sec:semantic}).

\begin{table}[t]
\caption{Adapting a cascade of ASR and LLM correction: weight space adaptation
fine-tuning FT) versus \vm{}. \vm{} runs forward passes only on the same
frozen endpoint used at deployment, so it needs no training infrastructure and
stays energy efficient on device. Figures are order-of-magnitude.}
\label{tab:vm-vs-training}
\centering
\small
\setlength{\tabcolsep}{5pt}
\renewcommand{\arraystretch}{1.25}
\begin{tabular}{@{}lcc@{}}
\toprule
 & \textbf{FT / LoRA} & \textbf{\vm{}} \\
\midrule
Adaptation pass   & forward + backward & \textbf{forward only} \\
Wall-clock        & hours+ & \textbf{minutes+} \\
Adapt data        & $10^3$ to $10^4$ utts & \textbf{$\sim$$10^2$ utts} \\
Model Params.     & $10^6$ to $10^9$ & \textbf{0} \\
Artifact size     & GB / MB & \textbf{$<$10\,KB} \\
\midrule
User Auditable    & \no & \yes \\
Portable          & \no & \yes \\
Reversible        & \no & \yes \\
\bottomrule
\end{tabular}
\end{table}

\section{The \vm{} Method}
\label{sec:method}

The corrector never changes; only its memory does. We formalize correction as a
per-utterance decision to act or abstain (\S\ref{sec:problem}), define the object we
optimize and the inference-only loop that optimizes it (\S\ref{sec:def}), and define a
measure of over-correction used in our analysis (\S\ref{sec:her}).

\subsection{Problem: correction against an oracle ceiling}
\label{sec:problem}

Let a domain provide $n$-best hypotheses $H = (h_1,\dots,h_n)$ from a frozen acoustic
model, with $h_1$ the 1-best. A corrector produces a single transcript from $H$. At low
error rates the corrector's main decision is not which hypothesis to pick but whether to
change $h_1$ at all, so we treat correction as a per-utterance choice to act or abstain.

The $n$-best list bounds what reranking can achieve. The oracle selects, per utterance,
the hypothesis in $H$ with the lowest WER against the reference, giving
$\mathrm{WER}_{\text{oracle}}$. Reranking cannot go below this bound; correction can,
because it may produce tokens that appear in no hypothesis. We measure how much of the
1-best-to-oracle gap a correction closes with the \emph{Recoverable Information Ratio}
(\rir{}):
\begin{equation}
\rho = \frac{\mathrm{WER}_{1\text{-best}} - \mathrm{WER}_{\hat y}}
            {\mathrm{WER}_{1\text{-best}} - \mathrm{WER}_{\text{oracle}}},
\label{eq:rho}
\end{equation}
where $\hat y$ is the corrected transcript. $\rho{=}1$ closes the gap exactly.
$\rho{<}0$ is the damage regime, where correcting is worse than keeping $h_1$.
$\rho{>}1$ means the corrector recovers tokens present in no hypothesis and goes below
the oracle bound. \rir{} is the target the method aims at, and \S\ref{sec:scaling}
reports it per domain.

To analyze \emph{why} a correction helps, we also label each token edit as helpful (fixed a wrong token) or harmful (broke a correct one); the Harmful Edit Rate (\her{}) is the harmful fraction, and isolates over-correction. We report \her{} on a subset of domains (Figure~\ref{fig:her}).

\subsection{Voice Memory and the optimization loop}
\label{sec:def}

A frozen corrector $M$ reads $H$ and an optional memory $s$, and emits
$\hat y = M(H, s)$. The memory $s$ is a short Markdown document, placed in the
corrector's context before each utterance. With $s = \varnothing$ this reduces to
standard GER. We keep $s$ in text, external to $M$, so that it can be read and edited by
an operator, reused across correctors, and added with no change to the inference path.
A learned $s$ is a small set of rules, for example:
\begin{quote}\ttfamily
keep the verbatim ASR token; do not normalize. \\
place `dollars' after the amount, not `\$' before it.
\end{quote}

We optimize $s$ on a domain's training split with a trajectory feedback loop
\cite{agrawal2025gepa,ni2026trace2skill,yang2026skillopt}, treating $s$ as the external state of the frozen $M$.
\begin{algorithm}[t]
\caption{Voice Memory optimization. All model calls are forward passes; no gradients, no weight updates.}
\label{alg:vm}
\DontPrintSemicolon
\SetAlgoNlRelativeSize{-1}
\footnotesize
\KwIn{frozen $M$, optimizer $Op$; splits $D_\text{train}, D_\text{sel}$; epochs $E$; budget $\{\ell_e\}$}
\KwOut{memory $s^\star$}
$s, s^\star \gets \varnothing$;\; $R \gets \emptyset$;\; $v^\star \gets \textsc{Score}(M, \varnothing, D_\text{sel})$\;
\For{$e \gets 1$ \KwTo $E$}{
  \ForEach{batch $H \subset D_\text{train}$}{
    $\hat{Y} \gets \{M(h, s) : h \in H\}$;\quad $F, U \gets$ failures, successes \tcp*{rollout}
    $s' \gets Op(s, F, U, R, \ell_e)$;\quad $v' \gets \textsc{Score}(M, s', D_\text{sel})$ \tcp*{$\le \ell_e$ edits}
    \eIf{$v' > v^\star$}{$s^\star \gets s \gets s'$;\; $v^\star \gets v'$ \tcp*{accept}}
    {$R \gets R \cup \{s'\}$ \tcp*{reject}}
  }
}
\Return $s^\star$
\end{algorithm}

The three splits stay disjoint: train supplies rollout evidence, selection gates
acceptance, and test is used once, for reporting. At deployment the frozen $M$ reads the
best accepted $s$. No weights change, and no extra model calls occur. The rules the loop
accepts are mostly suppressive; \S\ref{sec:scaling} shows the learned policy is
restraint.

\subsection{Measuring over-correction}
\label{sec:her}

WER reports whether a correction helped, not why. We add the \emph{Harmful Edit Rate}
(\her{}). We align $\hat y$ against $h_1$ and the reference, then label each token edit
as helpful (fixed a wrong token), harmful (broke a correct token), or missed. \her{} is
the fraction of all token edits that are harmful. It isolates over-correction: a
corrector that edits tokens that were already correct has a high \her{}.

We evaluate on HyPoradise v0 \cite{chen2023hyporadise}, which pairs Whisper 5-best
hypotheses with references across 10 domains under disjoint train/test splits: clean read
speech (\texttt{ls\_clean}), financial news (\texttt{wsj}), voice commands
(\texttt{atis}), telephone and meeting speech (\texttt{swbd}, \texttt{chime4}), and
accented or conversational speech (\texttt{cv}, \texttt{coraal}, \texttt{lrs2},
\texttt{td3}, \texttt{ls\_other}). A single frozen instruction LLM, the open-source MiniMax-M3~\cite{lai2026minimax} with 1M context and $\sim$428B parameters,
serves as both corrector and optimizer.\footnote{We use \texttt{MiniMax-M3}; the model checkpoint has fixed, open weights: \url{https://huggingface.co/MiniMaxAI/MiniMax-M3}.} Semantic scoring, used to gate some memories and analyzed in
\S\ref{sec:appendix-analyses}, runs on an on-device sentence encoder
(\texttt{all-MiniLM-L6-v2}, 4-bit; \cite{reimers2019sentencebert}) and adds no API
cost. We report the full agwer python suite, with WER shown alongside \rir{} and \her{}.\footnote{WER and \her{} are computed with our open-source \texttt{agwer} package (\url{https://pypi.org/project/agwer/} under MIT license), an efficient implementation that is safe to use as a reward target inside the optimization loop.}

\begin{table}[t]
\caption{\textsc{Voice Memory} (semantic-gated) across HyPoradise full test splits, same
corrector (\texttt{MiniMax-M3}$_{\text{428B}}$). \textbf{Bold} marks the lowest WER among
GER$_{\text{0-shot}}$, GER$_{\text{5-shot}}$, and \textsc{VM} per row. The gain follows the
scaling properties: $\rho_{\textsc{VM}}$ is high on high-WER \texttt{atis}/\texttt{cv} and negative
at the near-floor \texttt{td3}/\texttt{ls\_clean}. }
\label{tab:breadth}
\centering
\small
\begin{tabular}{l|cc|ccc|c}
\toprule
Domain & 1-best & Oracle & ger$_{\text{0}}$ & ger$_{\text{5}}$ & \textsc{VM}$_{\text{sem}}$ & $\rho_{\text{\textsc{VM}}}$ \\
\midrule
\texttt{atis}      & 7.9  & 4.7  & 6.3  & 5.2  & \cellcolor[HTML]{9AFF99} \textbf{4.9}  & \cellcolor[HTML]{9AFF99}\textbf{0.96} \\
\texttt{wsj}       & 4.9  & 3.6  & 5.8  & 4.7  & \textbf{4.3}  & 0.48 \\
\texttt{chime4}    & 9.9  & 7.5  & 9.7  & \textbf{9.4}  & \textbf{9.4}  & 0.21 \\
\texttt{cv}        & 15.4 & 11.3 & 13.1 & \textbf{12.7} & \cellcolor[HTML]{9AFF99}13.0 & \cellcolor[HTML]{9AFF99}\textbf{0.59} \\
\texttt{lrs2}      & 11.7 & 6.7  & 11.1 & \textbf{10.5} & 10.9 & 0.16 \\
\texttt{coraal}    & 25.6 & 23.7 & \textbf{25.1} & 25.9 & 25.3          & 0.15 \\
\texttt{td3}       & 4.1  & 2.9  & 4.4  & 4.3  & 4.2           & $-0.09$ \\
\texttt{ls\_o}& 3.7 & 2.3 & 4.1& 4.0 & \textbf{3.6} & 0.06   \\
\texttt{ls\_c}& 1.8 & 0.8 & 2.4& 2.5 & 1.9 & $-0.11$  \\
\bottomrule
\end{tabular}
\end{table}

\section{Over-Correction and Scaling Properties}
\label{sec:scaling}
\begin{figure}[t]
    \centering
    \includegraphics[width=0.55\linewidth]{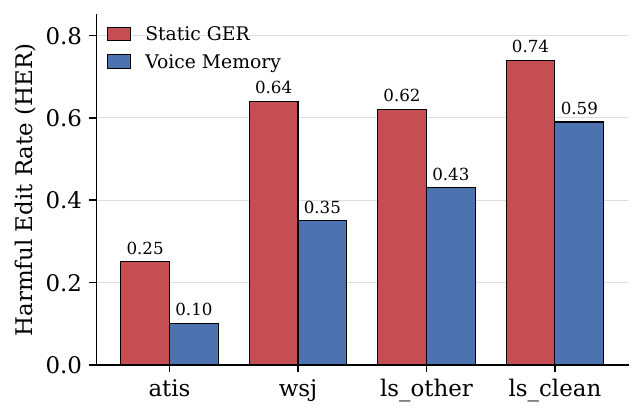}
    \caption{Harmful Edit Rate (HER) on four HyPoradise domains. Static GER over-corrects;
    \vm{} lowers HER on every domain. Lower is better.}
    \label{fig:her}
\end{figure}

\paragraph{Over-correction and restraint.}
The core failure of unconstrained correction is over-correction: zero-shot GER edits tokens that were already correct. In Table~\ref{tab:breadth} it raises WER above the 1-best on the low-headroom domains (\texttt{ls\_clean} $1.8\!\to\!2.4$, \texttt{ls\_other} $3.7\!\to\!4.1$, \texttt{td3} $4.1\!\to\!4.4$), and on four labeled domains it is harmful on a large fraction of its edits, up to 0.64 on \texttt{wsj} (Figure~\ref{fig:her}). \vm{} outperforms GER on four tasks. Moreover, it lowers the harmful-edit rate on every labeled domain (to 0.35 on \texttt{wsj}; Figure~\ref{fig:her}), and the deployed \texttt{wsj} memory is 776 bytes.

\paragraph{The gain scales with WER.}
One benefit of correction is that it scales with recoverable headroom, measured by
\rir{}: largest where 1-best WER is high, and vanishing at the error floor. Across the nine
domains of Table~\ref{tab:breadth}, \rir{} correlates with 1-best WER at $r{=}{+}0.90$.
\vm{} helps most where headroom is largest: at the \texttt{ls\_clean} floor
($\rho{=}{-}0.11$) and \texttt{td3} ($\rho{=}{-}0.09$) little remains to recover, while on
high-WER \texttt{atis} it recovers most of the gap ($\rho{=}0.96$). Few-shot prompting is
competitive on some domains but pays a token cost per call, whereas the \vm{} memory is
formed once and read for free.

\section{Memory Transfer and Portability}
\label{sec:transfer}
Since the memory is in text form, it is transferable to different models.
We show two additional studies: a memory written by one corrector helps a second corrector from a different family, and that second corrector can form its own memory with the same loop.

\begin{table}[t]
\caption{Robust \vm{} (VM) on the Robust HyPoradise corpora \cite{hu2024large}.
\vm{} lowers WER on the recorded-noise corpora (CHiME-4, NOIZEUS), where static GER
over-corrects, and is near-null on read speech with added noise (VoiceBank-DEMAND, LS-FreeSound; discussed in the text). $\rho_{\textsc{vm}}$ is the recovered fraction of the 1-best-to-oracle
gap; bold marks the best WER per row.}
\label{tab:robust}
\centering
\small
\setlength{\tabcolsep}{4pt}
\begin{tabular}{lrrrrr}
\toprule
Condition & 1-best & Oracle & Static & VM & $\rho_{\textsc{vm}}$ \\
\midrule
\multicolumn{6}{l}{\textit{CHiME-4} (real recorded / simulated noise)} \\
\quad test-real        & 9.8  & 7.8  & 9.9  & \textbf{9.5}  & 0.17 \\
\quad test-simu        & 12.7 & 10.3 & 12.5 & \cellcolor[HTML]{9AFF99}\textbf{11.7} & \cellcolor[HTML]{9AFF99}\textbf{0.39} \\
\quad dev-real         & 7.8  & 6.3  & 8.1  & \cellcolor[HTML]{9AFF99}\textbf{7.3}  & \cellcolor[HTML]{9AFF99}0.31 \\
\quad dev-simu         & 9.7  & 7.9  & 9.8  & \cellcolor[HTML]{9AFF99}\textbf{8.8}  & \cellcolor[HTML]{9AFF99}\textbf{0.47} \\
\midrule
\multicolumn{6}{l}{\textit{NOIZEUS, pooled by SNR}} \\
\quad 0\,dB            & 42.9 & 34.9 & 41.5 & \textbf{41.4} & 0.19 \\
\quad 5\,dB            & 14.5 & 9.2  & 14.8 & \cellcolor[HTML]{9AFF99}\textbf{12.6} & \cellcolor[HTML]{9AFF99}\textbf{0.36} \\
\quad 10\,dB           & 5.0  & 2.1  & 5.2  & \textbf{4.9}  & 0.05 \\
\quad 15\,dB           & 2.7  & 0.9  & 2.7  & 2.8  & $-0.06$ \\
\bottomrule
\end{tabular}
\end{table}

\begin{table}[t]
\caption{A second frozen corrector (Claude-4.6-Sonnet) at inference on a 100-utterance test,
varying only the memory. ``{+}MiniMax'' reads the memory optimized in \S\ref{sec:scaling}; ``{+}Claude'' reads a memory Claude formed itself with the same
validation-gated loop (\S\ref{sec:def}). On \texttt{atis} the self-formed memory
beats the MiniMax one and the $n$-best oracle ($\rho{=}1.28$); on data-hungry
\texttt{wsj} it over-corrects; at the low-WER floor no memory is best.}
\label{tab:claude}
\centering
\small
\setlength{\tabcolsep}{4.5pt}
\begin{tabular}{lrrrrr}
\toprule
Domain & 1-best & oracle & no mem & {+} mmax & {+} claude \\
\midrule
\texttt{atis}     & 8.2 & 4.9 & 6.7 & 6.1          & \textbf{3.9} \\
\texttt{wsj}      & 5.3 & 4.1 & 6.1 & \textbf{4.6} & 6.2 \\
\texttt{ls\_other}& 4.8 & 3.3 & \textbf{4.3} & 4.5 & 4.8 \\
\texttt{ls\_clean}& 1.9 & 0.7 & \textbf{1.7} & 1.8 & 1.8 \\
\bottomrule
\end{tabular}
\end{table}

\paragraph{A memory transfers to a corrector from a different family.}
We test a second frozen corrector, Anthropic Claude-4.6-Sonnet (2026 February), at inference. On \texttt{atis}, Claude reads a MiniMax-written memory and improves over its no-memory baseline (6.68\% to 6.08\% WER, Table~\ref{tab:claude}). The reader improves with a memory it did not write, so the memory is a separate component, not a property of the writer.

\paragraph{A corrector can form its own memory.}
Running the same validation-gated loop with Claude as both corrector and optimizer gives a self-formed memory that reaches $\rho{=}1.28$ on \texttt{atis}, above the MiniMax memory read by Claude and above the $n$-best oracle (3.94\% vs.\ 6.08\% WER,
Table~\ref{tab:claude}).

\paragraph{Transfer follows the scaling properties.}
The transfer gain concentrates where there is recoverable headroom, as
\S\ref{sec:scaling} predicts. Across the four domains of Table~\ref{tab:claude}, a
memory helps on high-WER \texttt{atis} but is neutral to slightly harmful at the clean
floor (\texttt{ls\_clean}, \texttt{ls\_other}), where little headroom remains. The learned rules are general restraint rather than a domain-specific fix, so the memory is reusable and can be applied to different models.

\begin{table}[t]
\caption{Voice Memory for X$\to$En translation correction (BLEU $\uparrow$). \vm{} improves over the 1-best in every direction ($\rho{>}0$); shown are the three directions where the formed memory also beats memory-free reconciliation (\texttt{fr}, where reconciliation alone is strongest, is discussed in Appendix~\ref{sec:appendix-translate}). $\rho$ is recoverable BLEU,
$(\textsc{vm}-\text{1-best})/(\text{oracle}-\text{1-best})$.}
\label{tab:translate}
\centering
\small
\setlength{\tabcolsep}{4.5pt}
\begin{tabular}{l|r|rrrr}
\toprule
Lang & 1-best & Oracle & zero-shot & \textsc{vm} & $\rho_{\textsc{vm}}$ \\
\midrule
\texttt{de} & 33.4 & 42.9 & 34.4 & \cellcolor[HTML]{9AFF99}\textbf{34.6} & 0.13 \\
\texttt{ja}  & 20.5 & 35.2 & 24.1 & \cellcolor[HTML]{9AFF99}\textbf{24.5} & 0.27 \\
\texttt{zh}   & 25.3 & 28.9 & 25.6 & \cellcolor[HTML]{9AFF99}\textbf{26.5} & 0.32 \\
\bottomrule
\end{tabular}
\end{table}

\paragraph{The mechanism generalizes beyond ASR.}
The recipe extends beyond speech recognition to speech-translation error correction (the GenTranslate setting \cite{hu2024gentranslate}), in which a corrector reconciles the $n$-best English-translation hypotheses of a non-English utterance and is scored by BLEU~\cite{li2024investigating}. In Table~\ref{tab:translate}, we run the identical \vm{} loop across four X$\to$En directions. \vm{} improves BLEU over the 1-best in every direction, recovering 13 to 32\% of the oracle headroom, and the optimizer learns the same restraint it learns for ASR: it declines to encode reference content present in no hypothesis, and rejects edits that would raise validation BLEU only by copying a single reference. A frozen corrector improving by editing a text memory is thus a general recipe for $n$-best language correction, not an ASR trick. This is a 40-utterance mini-study with a single corrector; the full setup is in Appendix~\ref{sec:appendix-translate}.

\paragraph{Memory provenance: quality beats coverage.}
Because the voice memory file is editable, its source matters more than its size. A memory generated by Claude Sonnet 4.6 reaches the best weighted WER at 7.17\%, ahead of the auto-generated memory (7.52), the expert hand-crafted memory (7.43), and the expert-seeded memory (7.32). The Claude memory is not the largest: at 5.4 KB it is smaller than the 9.0 KB auto-generated and 7.6 KB expert-seeded variants. Precise substitution entries and specific formatting rules govern correction quality more than raw coverage.

\section{Robust Voice Memory for Noisy ASR}
\label{sec:robust}

The restraint policy of \S\ref{sec:scaling} also holds under acoustic noise. RobustGER \cite{hu2024large} makes a corrector
noise-robust by distilling the $n$-best diversity into a language-space noise embedding and fine-tuning LLaMA against it, which costs one training run per deployment. \vm{} writes the same prior into a text file. We run the same pipeline on Whisper $n$-best from four noise families (CHiME-4, VoiceBank-DEMAND, LibriSpeech+FreeSound, NOIZEUS), optimize one semantic-gated memory per family, and read it at inference with no weight updates. Our VoiceBank-DEMAND 1-best WERs match RobustGER Table~1 to within $0.1$ point (party $22.59$ vs.\ $22.6$, baby $7.98$ vs.\ $8.0$), so the two setups are comparable.

\paragraph{The gain tracks how hard the speech is.}
As the scaling properties of \S\ref{sec:scaling} predict, \vm{} helps where 1-best WER and oracle headroom are large, and does nothing where they are small (Table~\ref{tab:robust}). It lowers WER on the genuinely degraded corpora, CHiME-4 and NOIZEUS, where static GER over-corrects. On
CHiME-4 dev-real, static GER raises WER ($7.75\!\to\!8.13$) while \vm{} lowers it
($7.75\!\to\!7.31$, $\rho{=}0.31$); on NOIZEUS~\cite{hu2007subjective} at $5$\,dB, \vm{} recovers
$14.51\!\to\!12.60$ ($\rho{=}0.36$; both in Table~\ref{tab:robust}). On VoiceBank-DEMAND
and LS-FreeSound, which are read speech with noise added rather than recorded, restraint
is already the right policy and the memory is near-null (party $\rho{=}0.03$). The result shows that the same restraint policy holds under noise: \vm{} matches RobustGER-style robustness on the degraded corpora with no fine-tuning and no latent noise vector, and correctly abstains where there is nothing to recover.

\section{Analysis: Meaning versus Surface Error}
\label{sec:appendix-analyses}
We consider that two questions sit underneath the scaling properties of \S\ref{sec:scaling}.
First, \emph{which} signal should the optimizer gate on: the metric we report
(i.e., WER and HER), or the quantity we actually care about (meaning)? Second, \emph{why}
would a meaning-based signal help at all? We answer both with measurements on
the same frozen corrector, and both answers reinforce the central finding that
good correction is mostly restraint.

\subsection{Meaning Is a Better Training Signal than WER}
\label{sec:semantic}

The validation gate in \S\ref{sec:def} can score candidates by any metric. We
compare two: a \textbf{WER gate} ($-\mathrm{WER}$ on the selection split) and a
\textbf{semantic gate} (mean cosine similarity between corrected and reference
embeddings from the on-device encoder). The two gates see the same rollouts, the
same candidate edits, and the same acceptance rule; they differ only in the
scalar they maximize, which isolates the effect of the objective itself. We
optimize both over two seeds and report on the full test split
(Table~\ref{tab:gate}).

\begin{table}[t]
\caption{Gate ablation, full test, two seeds. Optimizing the memory for
\emph{semantic} preservation beats optimizing it for WER \emph{on WER}, and is
$\sim$4$\times$ more seed-stable on \texttt{ls\_clean} (range 0.03 vs.\ 0.17).}
\label{tab:gate}
\centering
\small
\begin{tabular}{llrr}
\toprule
Domain & Gate & WER\% (mean) & WER range \\
\midrule
\multirow{2}{*}{\texttt{wsj}}
  & WER       & 4.35 & [4.17, 4.54] \\
  & Semantic  & \textbf{4.16} & [4.08, 4.23] \\
\midrule
\multirow{2}{*}{\texttt{ls\_clean}}
  & WER       & 2.16 & [2.08, 2.25] \\
  & Semantic  & \textbf{2.12} & [2.10, 2.13] \\
\bottomrule
\end{tabular}
\end{table}

The semantic gate wins on WER across both domains and both seeds, and resists
the optimization seed far better. The mechanism follows from
\S\ref{sec:scaling}: the optimal correction policy is mostly restraint, and a
semantic objective rewards restraint \emph{without} chasing the orthographic
ghosts (homophones, spelling variants) that a WER objective is tempted to
``\texttt{fix}.'' A WER gate, by construction, credits any edit that removes a surface
mismatch. This includes the meaning-preserving ones a careful reader would leave alone; the semantic gate withholds that credit, so fewer such edits survive acceptance. Optimizing the metric you report (\textit{i.e.}, WER) is thus \emph{worse}
than optimizing the quantity you care about (meaning), which WER only imperfectly
tracks. We frame this honestly: the margin is modest ($\sim$0.2 points), yet
consistent in direction across domains and seeds, and the stability and
meaning-preservation gains are unambiguous.

\subsection{Language Relativity: a Measurable Surface to Meaning Gap}
\label{sec:relativity}

We provide additional study inspired by \citet{hoijer1954sapir} on Whorf–Sapir hypothesis. A semantic objective can beat WER on WER because the two quantities live on
different axes, and we make this separation quantitative. For each utterance we measure a
\emph{surface} distance and a \emph{semantic} distance
($\cos$ of sentence embeddings). Regressing semantic on surface distance
yields a \emph{decoupling slope}; the fraction of surface-error mass whose
semantic distance falls below a small threshold is the \emph{benign mass}
(Table~\ref{tab:relativity}).

\begin{table}[t]
\caption{A measurable surface to meaning gap. The decoupling slope (0.38 to 0.60,
$\ll 1$) means a unit of WER moves meaning by only $\sim$half a unit; 53 to 68\%
of residual error \emph{mass} preserves meaning (semantic distance $< 0.15$).}
\label{tab:relativity}
\centering
\small
\begin{tabular}{lrrr}
\toprule
Domain & WER & Decoupling & Benign \\
       & (surf.) & slope & mass \\
\midrule
\texttt{atis}      & .080 & 0.60 & 66.4\% \\
\texttt{wsj}       & .046 & 0.38 & 67.5\% \\
\texttt{ls\_other} & .046 & 0.56 & 53.4\% \\
\texttt{ls\_clean} & .021 & 0.59 & 65.6\% \\
\bottomrule
\end{tabular}
\end{table}

The slope sits well below 1 in every domain: surface errors move meaning only
weakly. Most residual error mass is semantically benign. This measured by an independent on-device model, not asserted. This is an operational, falsifiable
statement of linguistic relativity for ASR: the text string and the underlying
message are distinct, and WER conflates them. It explains why over-correction is so costly: the corrector trades meaning-neutral surface forms for a lower apparent error count while leaving the message unchanged. It also explains why a semantic gate generalizes better. The practical consequence is a ceiling on how
much residual WER is worth pursuing: once an edit moves only the surface form and
not the meaning, chasing it trades risk of damage for no communicative gain.

\subsection{Error Analysis}
The residual after \vm{} splits into two buckets: genuine acoustic
confusions (named entities, rare words) that no language prior can resolve, and
annotation artifacts (spelling variants, spacing) that \emph{should not} be
``corrected.'' The first bucket is a limit of the cascade rather than of the
memory. The information needed to recover the token is simply absent from the
$n$-best list. The second is exactly where an unconstrained corrector does damage, and where the learned restraint pays off. The \her{}/\rir{} decomposition (\S\ref{sec:her}) separates
these, and the benign-mass measure (\S\ref{sec:relativity}) bounds how much of
the remaining WER is worth chasing at all. Appendix~\ref{sec:cases} gives eight representative such cases.

\section{Conclusion}
\label{sec:conclusion}
\vm{} makes a frozen corrector better without touching a single weight. Everything the
system learns lives in a small, auditable text memory read at inference time; the only
difference between a worse corrector and a better one is that file. A better agent here
is simply a better text file. The policy the file comes to encode is restraint, knowing
which tokens to leave alone, and this is why we cast correction as a per-utterance
decision to act or abstain rather than as a sequence-to-sequence rewrite, with a
listener that decides at stream time and a thinker that revises the memory
asynchronously.

The pieces compose into one picture. The gain scales with recoverable headroom: the
same policy that repairs far-field speech holds still on clean read speech. A memory
written by one model family improves a reader from another, because explicit rules
travel where weights cannot; the same trust can be trained into weights from audio
\cite{wan2026speechhands}, but \vm{} acquires it from inference-time feedback, with backpropagation. The cheapest adaptation and personalization we know of for an ASR-LLM
cascade. We would refocus the task
itself: optimize not orthography but meaning, and decide not how often to act but when
to hold still~\cite{mullennix2011typicality, sheffert1995effects}.

\section*{Limitations}
Our main analyses use a single open-source model with 428B parameters and 23B activated parameters, \texttt{MiniMax-M3}.
Appendix~\ref{sec:appendix-qwen} reproduces the core effect with an open
Qwen3-30B-A3B corrector across ten datasets.
 \rir{} and the benign-mass
measure depend on the sentence encoder we use. We reduce this dependence with a small,
public, on-device model with an average system latency of 610\,ms, and by releasing the protocol and audio-visual memory formation~\cite{kim2026two, ghosh2024lipger}.

\section*{Ethics Statement}
\vm{} produces a human-readable artifact, which makes auditing and bias review easier
than inspecting weight updates. Domain memories can encode dataset-specific conventions,
such as spelling norms, that should not transfer across dialects or populations without
review. The benign-mass analysis is one way to surface these conventions before
deployment.

\bibliography{custom}
\bibliographystyle{iclr2026_conference}

\clearpage
\appendix

\section{Reproducibility and Hyperparameters}
\label{sec:appendix}
We default to 4 epochs, rollout batch 24, selection split 80, and textual
learning rate 4 with cosine decay (floor 2), under the ``strictly
greater'' validation gate. Each domain uses three disjoint splits, and we
report test results once. All semantic scoring runs on-device.

\section{Broad Study with an Open Corrector}
\label{sec:appendix-qwen}

The main body establishes \vm{} in depth with the MiniMax-M3 corrector. The effect is not an artifact of one model: an open, inference-only corrector reproduces it at full breadth. We pair Whisper-v2-Large~\cite{radford2023robust} as the frozen decoder with Qwen3-30B-A3B~\cite{qwen2025qwen3}. A sparse Mixture-of-Experts that activates only $\approx$3B of its 30B parameters per forward pass (i.e., as the corrector.) The agent reads a per-domain \texttt{memory.md} at inference time and never updates a weight. Across 16,108 test samples from 10 datasets, this open-corrector breadth study complements the main-body MiniMax-M3 depth study.

\begin{table}[th!]
\caption{Main results for the open corrector: full test set WER (\%) across 10 benchmarks. Raw ASR is the Whisper-v2-Large 1-best; \vm{} (All Combined) applies negative rules, filtered memory, domain prompts, and oracle gating; Few-shot adds three in-context correction examples per dataset. The \textbf{WEIGHTED} row is a single global error ratio across all datasets.}
\label{tab:qwen-main}
\centering
\small
\begin{tabular}{@{}lrrr@{}}
\toprule
\textbf{Dataset} & \textbf{Raw ASR WER} & \textbf{VM (Combined)} & \textbf{Few-shot} \\
\midrule
\texttt{ATIS}       & 8.40  & 3.40  & 3.00  \\
\texttt{CHiME-4}    & 12.69 & 10.46 & 10.28 \\
\texttt{CORAAL}     & 25.83 & 25.65 & 25.44 \\
\texttt{CV}         & 15.44 & 13.16 & 13.16 \\
\texttt{LRS2}       & 11.85 & 10.69 & 10.78 \\
\texttt{LS-Clean}   & 1.79  & 1.63  & 1.64  \\
\texttt{LS-Other}   & 3.67  & 3.39  & 3.41  \\
\texttt{SWBD}       & 15.86 & 15.47 & 15.33 \\
\texttt{TED-3}      & 4.44  & 4.20  & 4.05  \\
\texttt{WSJ}        & 6.03  & 5.38  & 5.36  \\
\midrule
\textbf{WEIGHTED}   & \textbf{8.36} & \textbf{7.52} & \textbf{7.47} \\
\bottomrule
\end{tabular}
\end{table}

Table~\ref{tab:qwen-main} confirms the depth-study pattern at breadth: \vm{} cuts weighted WER from 8.36 to 7.47, and no dataset regresses below its raw baseline. Gains concentrate where memory supplies domain vocabulary the corrector cannot otherwise reach the performance. \texttt{ATIS} falls from 8.40 to 3.00 and \texttt{CHiME-4} from 12.69 to 10.28, while acoustically ambiguous conversational speech (\texttt{CORAAL}, \texttt{SWBD}) moves little.

The component ablation isolates which interventions earn that gain, and the answer is restraint. Negative rules are the single load-bearing component: alone, they drive weighted WER from 7.66 to 7.52, more than any other fix. Their job is to stop the corrector from rewriting tokens that were already correct. This is suppressing contraction edits, paraphrase, and digit-word conversion, so the agent's discovered policy is largely knowing when \emph{not} to act. Few-shot examples add a further 0.05 on top, reaching 7.47. Removing the memory entirely regresses the system to 7.66, the level of having no negative rules at all, which quantifies the memory's contribution as a knowledge source the corrector's parametric weights do not supply. A better agent here is simply a better text file.

\paragraph{The irreducible error floor.}
Residual WER does not imply residual headroom. We quantify a \emph{tolerated WER} by forgiving two error classes that no text-only corrector can resolve without the source document: out-of-vocabulary proper nouns and British/American spelling variants. On \texttt{LS-Clean}, 42.9\% of residual errors are forgivable under these two rules, and accounting for structurally irreducible errors yields a practical hard floor of 0.749\%. Most of what \vm{} leaves uncorrected on clean read speech is unreachable in principle, not a failure of the agent's policy.

\section{Corrector-Agnostic Memory: a Claude Case Study}
\label{sec:appendix-claude}

Because the memory is text, it need not be tied to the corrector that wrote it.
We test this with a second, independent frozen corrector applied
purely at inference (no fine-tuning) on \texttt{atis}, \texttt{wsj},
\texttt{ls\_other}, and \texttt{ls\_clean}. These runs are small-sample and
illustrative (50 to 100 utterances); we read them through \rir{}, which is sample-internal, and
leave a full-test, same-corrector comparison to future work.

Three findings emerge (Table~\ref{tab:claude}). \textbf{(i) The skill ports across
correctors.} A memory written by MiniMax lifts a different, stronger frozen reader
it never trained with, and that reader \emph{out-restrains} the writer at the clean
floor, with no over-correction. \textbf{(ii) The corrector can form its own memory.}
Running the validation-gated loop of \S\ref{sec:def} with Claude as both corrector
and optimizer yields, on \texttt{atis}, a memory that beats the MiniMax-written one
read by the same Claude ($3.94$ vs.\ $6.08$, $\rho{=}1.28$, past the oracle). For instance, no
privileged optimizer is required when the domain has learnable patterns and
headroom. \textbf{(iii) Transfer is roughly neutral.} Applying each memory across
the four domains (a $4{\times}4$ matrix) leaves WER close to the own-memory
diagonal: the learned rules are general restraint, not domain overfits. The
boundaries are honest that self-formation loses on number-formatting-heavy \texttt{wsj}
(data-hungry) and at the LibriSpeech floor (nothing to learn) but the headline
holds: the memory is the portable, corrector-agnostic asset, and a strong reader can
re-form a better one where the domain affords it.

\section{X$\to$En Voice Memory: a Translation Mini-Study}
\label{sec:appendix-translate}

Is the agentic memory mechanism specific to ASR, or does it carry to other
$n$-best language tasks? We test it on \emph{speech translation} error correction,
the GenTranslate setting \cite{hu2024gentranslate, imai2025evaluating}: a corrector reconciles the
$n$-best English-translation hypotheses of a non-English source utterance into a
single better translation, scored by \textbf{BLEU}. We run the identical Voice
Memory loop (\S\ref{sec:def}) a fixed LM as both corrector and BLEU-gated
optimizer on four X$\to$En directions spanning three language families: \texttt{de}, \texttt{ja} (CoVoST-2 speech translation) and
\texttt{zh} (FLEURS machine translation; CoVoST-2 has no Chinese and FLEURS no
French, so we take the best available source per language). Each memory is formed
on 10 rollout + 20 select items and read on a held-out 40-utterance test.

Table~\ref{tab:translate} shows the mechanism transfers. \vm{} improves BLEU over
the 1-best in every direction, recovering $13$ to $32\%$ of the oracle headroom, and
the formed memory adds over memory-free reconciliation on three of four languages.
The familiar boundary reappears on \texttt{fr}, where Claude's reconciliation alone
is already strong (53.1) and the memory slightly over-applies (51.1) to restraint
again. Most telling are the formation traces: the optimizer repeatedly identified
reference content present in \emph{no} hypothesis (place names, idioms, legal terms)
as unrecoverable and declined to encode guesses, and it \emph{rejected} candidate
edits that would have raised select BLEU only by copying a single reference. The
same enforced-restraint discipline the ASR memories learn. A frozen corrector
improving by editing an auditable text skill is thus not an ASR trick but a general
recipe for $n$-best language correction. This is a 40-utterance mini-study with a
single corrector; we leave full-scale, multi-corrector translation to future work.

\section{Qualitative Case Studies}
\label{sec:cases}
We show eight representative cases (Table~\ref{tab:cases}) spanning the
\her{}/\rir{} taxonomy: \vm{} fixes genuine errors, suppresses the harmful edits
static GER introduces, and in the generative cases recovers a transcript present in
no hypothesis.

\begin{table}[t]
\caption{Qualitative case studies across four ASR domains. Each block shows the Whisper 1-best, the static GER correction, the \vm{} correction, and the gold reference (verbatim, lowercased as scored). The bracketed tag is the \vm{} edit category. \textbf{Generative} cases show \vm{} producing a transcript absent from every n-best hypothesis (e.g., recovering the spelled-out form \texttt{s b} / \texttt{b n} from \texttt{sb} / \texttt{be in}). \textbf{Harmful-suppressed} cases show static GER over-correcting a clean 1-best (introducing \texttt{me l}, normalizing \texttt{\$66.6}, or deleting \texttt{words}) while \vm{} correctly holds back. \textbf{Helpful} cases show \vm{} repairing a genuine error and beating the 1-best WER.}
\label{tab:cases}
\small
\centering
\setlength{\tabcolsep}{4pt}
\begin{tabular}{@{}llp{0.24\textwidth}p{0.24\textwidth}p{0.24\textwidth}@{}}
\toprule
\textbf{Domain} & \textbf{Field} & \textbf{1-best} & \textbf{Static GER} & \textbf{\vm{}} \\
\midrule

\multirow{2}{*}{ATIS}
& transcript & \texttt{what is meal code sb} & \texttt{what is me l code sb} & \texttt{what is meal code s b} \\
& \multicolumn{4}{@{}p{0.82\textwidth}@{}}{\raggedright Reference: \texttt{what is meal code s b} \quad\textbf{[helpful; generative]}} \\
\addlinespace

\multirow{2}{*}{ATIS}
& transcript & \texttt{what does fair code be in mean} & \texttt{what does faircode be in mean} & \texttt{what does fare code b n mean} \\
& \multicolumn{4}{@{}p{0.82\textwidth}@{}}{\raggedright Reference: \texttt{what does fare code b n mean} \quad\textbf{[helpful; generative]}} \\
\addlinespace

\multirow{2}{*}{ATIS}
& transcript & \texttt{i need to flight from oakland to salt lake city on wednesday departing after six p m} & \texttt{i need to fly from oakland to salt lake city on wednesday departing after six p m} & \texttt{i need a flight from oakland to salt lake city on wednesday departing after six p m} \\
& \multicolumn{4}{@{}p{0.82\textwidth}@{}}{\raggedright Reference: \texttt{i need a flight from oakland to salt lake city on wednesday departing after six p m} \quad\textbf{[helpful; harmful-suppressed]}} \\
\addlinespace

\multirow{2}{*}{ATIS}
& transcript & \texttt{show me run trip fares from toronto to detroit on delta northwest u s air and united airlines} & \texttt{show me round trip fares from toronto to detroit on delta northwest us air and united airlines} & \texttt{show me round trip fares from toronto to detroit on delta northwest u s air and united airlines} \\
& \multicolumn{4}{@{}p{0.82\textwidth}@{}}{\raggedright Reference: \texttt{show me round trip fares from toronto to detroit on delta northwest u s air and united airlines} \quad\textbf{[helpful]}} \\
\addlinespace

\multirow{2}{*}{WSJ}
& transcript & \texttt{kyster tech did not name successors for mister mayer in his post as the parent} & \texttt{kyster tech did not name a successor for mister mayer in his post as the parent} & \texttt{kyster tech did not name successors for mister mayer in his post at the parent} \\
& \multicolumn{4}{@{}p{0.82\textwidth}@{}}{\raggedright Reference: \texttt{kaisertech did not name successors for mister maier in his posts at the parent} \quad\textbf{[helpful]}} \\
\addlinespace

\multirow{2}{*}{WSJ}
& transcript & \texttt{the bid for the remaining shares is valued at \$sixty-six point six million} & \texttt{the bid for the remaining shares is valued at \$66.6 million} & \texttt{the bid for the remaining shares is valued at sixty six point six million} \\
& \multicolumn{4}{@{}p{0.82\textwidth}@{}}{\raggedright Reference: \texttt{the bid for the remaining shares is valued at sixty six point six million} \quad\textbf{[no\_edit; harmful-suppressed]}} \\
\addlinespace

\multirow{2}{*}{LS-clean}
& transcript & \texttt{words was it their colours} & \texttt{was it their colours} & \texttt{words was it their colours} \\
& \multicolumn{4}{@{}p{0.82\textwidth}@{}}{\raggedright Reference: \texttt{words was it their colours} \quad\textbf{[no\_edit; harmful-suppressed]}} \\
\addlinespace

\multirow{2}{*}{LS-other}
& transcript & \texttt{bessy do you wish to die} & \texttt{bessie do you wish to die} & \texttt{bessy do you wish to die} \\
& \multicolumn{4}{@{}p{0.82\textwidth}@{}}{\raggedright Reference: \texttt{bessy do you wish to die} \quad\textbf{[no\_edit; harmful-suppressed]}} \\

\bottomrule
\end{tabular}
\end{table}

\end{document}